\documentclass{article}

\usepackage{iclr2026_conference,times}
\iclrfinalcopy
\usepackage{amsmath}
\usepackage[utf8]{inputenc} 
\usepackage[T1]{fontenc}    
\usepackage{hyperref}       
\usepackage{url}            
\usepackage{graphicx}       
\usepackage{booktabs}       
\usepackage{amsfonts}       
\usepackage{nicefrac}       
\usepackage{microtype}      
\usepackage{xcolor}         
\usepackage{pgfplots}       
\pgfplotsset{compat=1.18}
\usepackage{multicol,multirow}
\usepackage{svg}
\usepackage{enumitem}
\usepackage{caption}
\usepackage{placeins} 
\usepackage{float}
\usepackage{cuted}
\usepackage{hyperref}

\definecolor{citecolor}{HTML}{0071bc}
\definecolor{revisionred}{HTML}{CC0000}
\newif\ifrevisioncolor
\revisioncolortrue

\definecolor{paleplum}{rgb}{0.8, 0.6, 0.8}
\hypersetup{
  colorlinks,
  citecolor=citecolor,
  linkcolor=red
}
\title{FlashAR: Efficient Post-Training Acceleration for Autoregressive Image Generation}

\author{%
    \parbox{\textwidth}{\centering
        \textbf{Junkang Zhou}\textsuperscript{1}$^*$ ~~
        \textbf{Yefei He}\textsuperscript{1}$^{*\dagger\ddagger}$ ~~
        \textbf{Feng Chen\textsuperscript{2}$^{*\dagger}$} ~~
        \textbf{Weijie Wang\textsuperscript{1}} ~~
        \textbf{Bohan Zhuang\textsuperscript{1}$^\ddagger$}
    }\\[6pt]
    \parbox{\textwidth}{\centering
        \textsuperscript{1} Zhejiang University, China \hspace{1.5em}
    }\\
    \parbox{\textwidth}{\centering
        \textsuperscript{2} University of Adelaide, Australia
    }\\[4pt]
    \parbox{\textwidth}{\centering
        \footnotesize
        $^{\ast}$ Equal contribution \quad
        $\dagger$ Project lead \quad
        $^{\ddagger}$ Corresponding authors
    }
}

\begin{document}

\maketitle

{
\maketitle
\begin{center}
    \centering
    \includegraphics[width=1.0\textwidth]{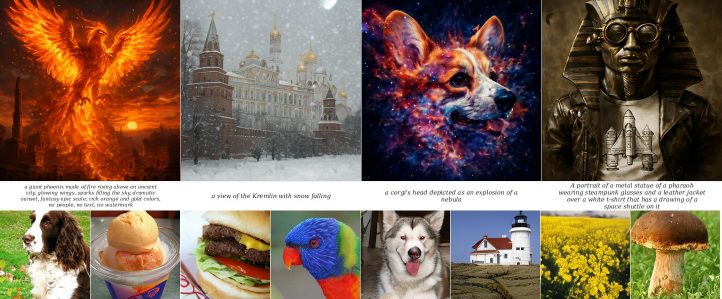}
    \captionof{figure}{\textbf{Generated samples from FlashAR.} The first row shows $512\times512$ text-guided generation results, while the second row presents class-conditional generation samples at $384\times384$ and $256\times256$ resolutions.}
\end{center}
}

\begin{abstract}
Large-scale autoregressive models have demonstrated remarkable capabilities in image generation. However, their sequential raster-scan decoding relies on strictly next-token prediction, making inference prohibitively expensive. 
Existing acceleration methods typically either introduce entirely new generation paradigms that necessitate costly pre-training from scratch, or enable parallel generation at the expense of a training-inference gap or altered prediction objectives.
In this paper, we introduce \textbf{FlashAR}, a lightweight post-training adaptation framework that efficiently adapts a pre-trained raster-scan autoregressive model into a highly parallel generator based on two-way next-token prediction. 
Our key insight is that effective adaptation should minimize modifications to the pre-trained model’s original training objective to preserve its learned prior.
Accordingly, we retain the original AR head as a horizontal head for row-wise prediction and introduce a complementary, lightweight vertical head for column-wise prediction.
To facilitate efficient adaptation, we branch the vertical head from a pre-final layer rather than the final layer, bypassing the inherent horizontal head bias.
Moreover, since horizontal and vertical predictions capture complementary dependencies whose relative importance varies across target positions, we employ a learnable fusion gate to dynamically combine the two predictions at each position.
To further reduce adaptation cost, we propose a two-stage adaptation pipeline: the vertical head is first initialized through adaptation from the pre-trained autoregressive model before jointly fine-tuned with backbone to adapt to the new decoding paradigm.
Extensive experiments on LlamaGen, GLM-Image and Emu3.5-Image show that FlashAR achieves a \textbf{13.03$\times$} speedup for $512\times512$ generation using approximately \textbf{0.05\%} of its pre-training tokens. Our code is available at \href{https://lxazjk.github.io/FlashAR/}{here}.
\end{abstract}

\section{Introduction}
Autoregressive (AR) models~\cite{cui2025emu35,emu3,glmimage2026,liu2024lumina,xin2025lumina2,team2025nextstep,wang2025simplear,geng2025xomni} have emerged as a powerful paradigm for high-fidelity image generation. By representing visual data as discrete token sequences~\cite{oord2018neuraldiscreterepresentationlearning,esser2021tamingtransformershighresolutionimage}, these models achieve strong scalability and generation quality. However, these models remain fundamentally constrained by raster-scan decoding, which generates image tokens sequentially from left to right and top to bottom. As a result, the decoding latency grows linearly with the number of image tokens, making high-resolution generation prohibitively slow. Moreover, predicting only one token at each step prevents AR decoding from effectively utilizing the parallel computing capabilities of modern GPUs.

Existing acceleration methods have attempted to mitigate this bottleneck, but they often introduce new limitations. One line of work redesigns the generation paradigm, for example by changing the token prediction order~\cite{nar,wang2025par,zhang2025locality,li2025randomizedAR,yu2025randomized} or adopting multi-scale autoregressive generation~\cite{tian2024var,han2025infinity,tang2024hart}. Although such methods can substantially reduce decoding latency, they usually require costly pre-training from scratch and cannot directly reuse the large repository of existing pre-trained raster-scan AR models. Another line of work enables parallel generation through post-training adaptation, such as discrete diffusion adaptation~\cite{cui2025emu35}. However, these approaches modify the original prediction objective and introduce a discrepancy between pre-training and inference. This dilemma raises a critical open question: \emph{Can we transform a pre-trained raster-AR model into a highly parallel generator while inheriting its powerful generative capabilities and keeping the post-training overhead minimal?}

\begin{figure*}
    \centering
    \includegraphics[width=\linewidth]{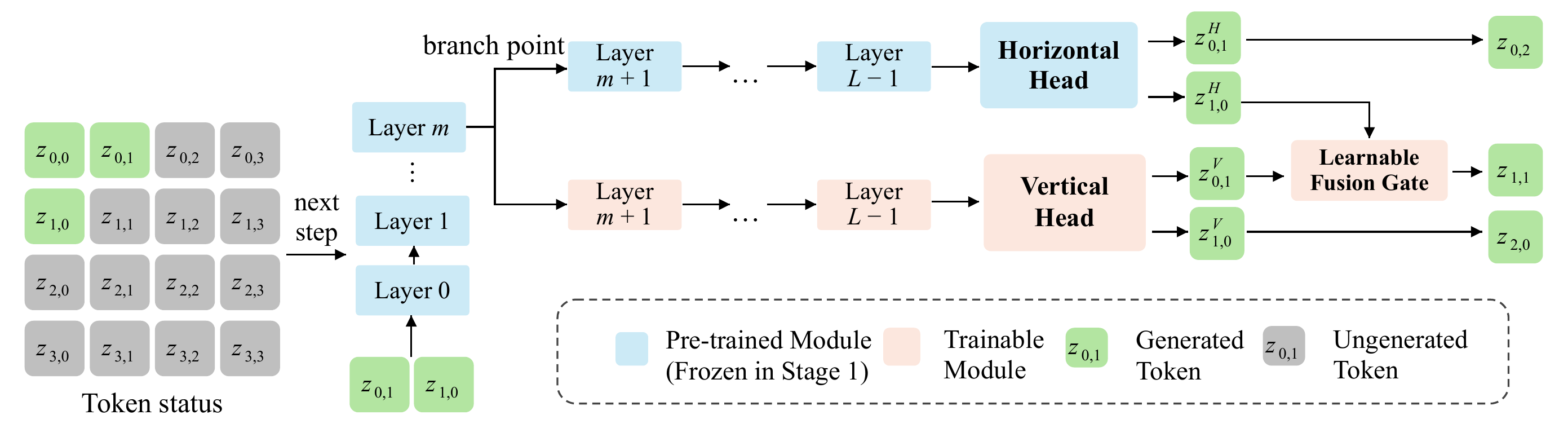}
    \caption{Overview of the FlashAR framework. Initialized from a pre-trained raster-scan autoregressive model, the architecture incorporates a pre-final branching module for the vertical head and a learnable fusion gate, facilitating efficient adaptation to parallel generation.}
    \label{fig:overview}
\end{figure*}

To address these challenges, we propose \textbf{FlashAR}, a lightweight post-training adaptation framework that efficiently adapts a pre-trained raster-scan autoregressive model into a highly parallel generator by reusing its horizontal predictor and adding vertical and fused prediction objectives. As shown in Figure~\ref{fig:overview}, we retain the original AR head as a horizontal head for row-wise prediction and introduce a lightweight vertical head for column-wise prediction, which enables the model to support parallel decoding under a two-way next-token prediction structure.

A key challenge in introducing vertical prediction is that the pre-trained model is inherently biased toward the original horizontal raster-scan objective and directly attaching a vertical head to the final layer is therefore suboptimal. To facilitate efficient adaptation and bypass this inherent horizontal head bias, we branch the vertical head from a pre-final layer rather than from the final layer, allowing the vertical pathway to access richer representations.

Moreover, horizontal and vertical predictions capture complementary directional dependencies, and their relative importance varies across spatial positions. A fixed combination of the two predictions is therefore insufficient and may introduce spatial inconsistency or context conflicts. To address this issue, FlashAR employs a learnable fusion gate that dynamically combines horizontal and vertical predictions at each target position, allowing the model to exploit their complementarity.
To further reduce adaptation cost, we propose a two-stage post-training pipeline. The newly introduced vertical head is first initialized through adaptation from the pre-trained autoregressive model, allowing it to rapidly acquire meaningful prediction ability. In the second stage, the vertical head and backbone are jointly fine-tuned to adapt the model to the new two-way decoding paradigm. 
Finally, to translate this theoretical parallelism into inference speedups, we deploy a hardware-aware inference pipeline. By leveraging FlexAttention to dynamically compile highly sparse 2D proximity masks on the fly, coupled with batched KV-cache updates, FlashAR alleviates memory and kernel launch bottlenecks, thereby improving practical inference speed.
%
Extensive experiments on LlamaGen, GLM-Image, and Emu3.5 demonstrate that FlashAR achieves a 13.03$\times$ speedup with negligible performance decline for $512\times512$ generation using approximately 0.05\% of its pre-training tokens.

In summary, our main contributions are as follows:

\begin{itemize} [leftmargin=*]

\item We propose \textbf{FlashAR}, a lightweight post-training acceleration framework that efficiently transforms pre-trained raster-scan AR models into highly parallel generators, largely retaining their generative capabilities without the prohibitive cost of training from scratch.


\item  We introduce a dual-head branching architecture with a two-stage post-training pipeline to resolve the tension between preserving a pre-trained model's horizontal prior and introducing orthogonal vertical prediction. 


\item We demonstrate the effectiveness of FlashAR across both class-conditional and text-to-image models, achieving a 13.03$\times$ speedup on Emu3.5-Image with limited post-training cost and competitive generation quality.
\end{itemize}

\section{Related Work}

\subsection{Autoregressive Visual Generation}

Building on the remarkable success of large language models (LLMs)~\cite{touvron2023llama,bai2023qwen,yang2025qwen3,team2023gemini}, autoregressive (AR) architectures have rapidly emerged as a compelling paradigm for visual and multimodal generation \cite{duan2026liveworld}. 
These approaches typically rely on learned visual tokenizers to quantize continuous images into discrete latent tokens. By flattening these 2D token maps into 1D sequences, visual generation is naturally cast as a next-token prediction task under a strict causal objective. Early pioneering works, including PixelCNN~\cite{van2016pixelcnn}, iGPT~\cite{chen2020generative} and Parti~\cite{yu2022parti}, demonstrated the strong potential of this formulation. More recently, driven by well-established scaling laws, large-scale models---such as Emu~\cite{emu3,cui2025emu35}, NextStep-1~\cite{team2025nextstep}, Lumina-mGPT~\cite{liu2024lumina,xin2025lumina2} and GLM-Image~\cite{glmimage2026}---have substantially advanced the paradigm. By unifying visual modeling under a rigorous causal objective, these models effectively capture complex long-range spatial dependencies, achieving quality that rivals or surpasses contemporary diffusion-based approaches.

Despite these advantages, standard AR models for visual generation remain fundamentally bottlenecked by their reliance on 1D raster-scan decoding. Generating tokens strictly sequentially---left to right, top to bottom---introduces serial dependencies that scale quadratically with sequence length. For high-resolution image synthesis, this requires executing thousands of sequential forward passes per image. Consequently, the decoding process becomes heavily memory-bandwidth-bound, resulting in inference latencies that are prohibitive for real-time or interactive applications.

\subsection{Efficient and Parallel Autoregressive Decoding}

To mitigate the inference latency of standard visual AR models, recent work has explored several parallelization strategies. One line of research accelerates decoding by departing from strict raster-scan token ordering. For instance, VAR~\cite{tian2024var,han2025infinity,tang2024hart} reformulates generation as coarse-to-fine ``next-scale prediction'', while NAR~\cite{nar} exploits spatial locality through ``next-neighbor prediction''. PAR~\cite{wang2025par} partitions image tokens into subsets and applies the standard next-token prediction paradigm within each subset. Although these structural changes successfully reduce sequential dependencies, they require bidirectional attention mechanisms and thus must be trained from scratch---making them fundamentally incompatible with existing pre-trained raster-AR models.
A second line bridges AR and diffusion frameworks via discrete diffusion adaptation~\cite{deng2025uniform,gat2024discrete,arriola2025blockdiffusion,shi2025muddit}. Methods such as Emu3.5~\cite{cui2025emu35} replace lengthy AR chains with parallel refinement over noisy token blocks. While this improves throughput, imposing a diffusion objective fundamentally alters the causal structure of the AR backbone, necessitating multi-stage fine-tuning and often compromising the fine-grained structural consistency that strict causal modeling naturally provides.
A third alternative involves speculative decoding~\cite{teng2024SJD,jang2024lantern,wang2024continuousSD}, which serves as a training-free acceleration plug-in. However, the practical speedup of speculative decoding is fundamentally constrained by the acceptance rate of the draft model, typically yielding marginal gains compared to architectural parallelization. Therefore, we do not include speculative decoding as a primary baseline for comparison.

In contrast, FlashAR focuses on accelerating existing raster-scan AR models through lightweight post-training. Building on the two-direction decoding of NAR~\cite{nar}, it introduces a pre-final vertical branch and a learnable fusion gate to enable parallel generation while retaining the pre-trained horizontal predictor.

\section{Preliminaries}

\subsection{Standard Raster-Scan Autoregressive Image Generation}

Given a generation condition $c$ and a discrete image token grid $Y=\{y_{p,q}\}_{p=0,q=0}^{H-1,W-1}$ of size $H \times W$, standard autoregressive image generation~\cite{llamagen, emu3} flattens the 2D token grid into a 1D sequence using raster-scan order. The conditional distribution is factorized as
\begin{equation}
p(Y \mid c)=\prod_{i=0}^{HW-1} p(y_i \mid y_{<i}, c).
\end{equation}

While this formulation is expressive, it enforces strictly sequential decoding over all $HW$ image tokens. As a result, the inference cost grows proportionally with the total number of visual tokens, which becomes increasingly expensive for high-resolution image synthesis.

\subsection{Diagonal-Step Factorization}

Following NAR~\cite{nar}, we factorize the autoregressive image generation into two orthogonal directions, where the original decoding head serves as next-token predictor for the row-wise prediction and an additional vertical head is introduced for column-wise prediction. Consequently, it partitions the 2D grids into
\begin{equation}
\mathcal{D}_t=\{y_{p,q}\mid p+q=t\}, \qquad t=0,1,\dots,H+W-2.
\end{equation}

This yields a diagonal-step factorization of the image distribution:
\begin{equation}
p(Y \mid c)=\prod_{t=0}^{H+W-2} p(\mathcal{D}_t \mid \mathcal{D}_{<t}, c).
\end{equation}

Under this formulation, decoding proceeds across diagonal steps rather than individual tokens, reducing the number of sequential iterations from $HW$ to $H+W-1$. Therefore, the generation process changes from quadratic complexity over the 2D token grid to linear complexity in the spatial dimensions, enabling substantially more efficient autoregressive image synthesis.

\section{Methodology}
To avoid the prohibitive training costs of native parallel paradigms, our goal is to adapt a pre-trained visual AR model into a highly parallel two-way generator while aiming to retain its generative capabilities. Instead of retraining the backbone from scratch~\cite{nar}, we propose \textbf{FlashAR}, an elegant and lightweight post-training adaptation framework. This approach reuses the pre-trained causal decoder, introduces an additional vertical branch at a pre-final layer, and seamlessly fuses the predictions of orthogonal heads through a dynamic gating mechanism.

\begin{figure*}
    \centering
    \includegraphics[width=\linewidth]{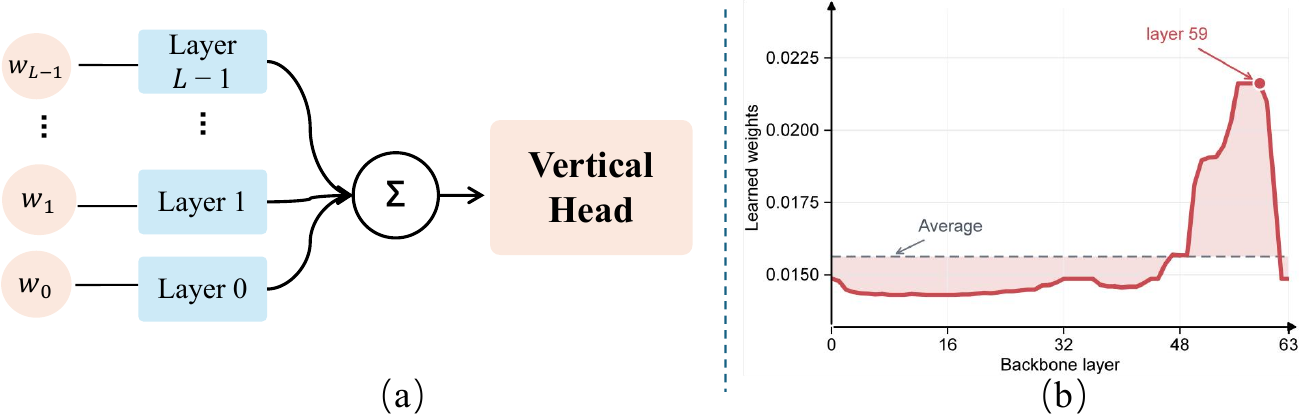}
    \caption{Analysis of linear probing experiments. (a) Schematic illustrating the aggregation of features from all Transformer layers using learnable weights before input to the vertical head. (b) The probe assigns greater weight to pre-final features than to the final-layer representation.}
    \label{fig:linear_probe}
\end{figure*}


\subsection{Pre-final Branching for Dual-head Decoding}
Recent studies~\cite{wu2024linguistic,skean2025layer} suggest that top-layer representations in deep autoregressive models can become increasingly specialized to the final prediction objective, losing general semantic abstractions. Moreover, the empirical observation illustrated in Figure~\ref{fig:linear_probe}  suggests that top-layer representations in a pre-trained raster-scan autoregressive model become specialized to the original left-to-right next-token prediction objective. Therefore, instead of introducing the new prediction branch at the final layer~\cite{nar}, we branch from a pre-final layer of the decoder, where the features remain semantically rich and are less tightly coupled to the original decoding direction.

For the probing experiment in Figure~\ref{fig:linear_probe}, we freeze the backbone and jointly train the layer weights and vertical prediction head using the vertical next-token loss.

Let the pre-trained decoder backbone consist of $L$ transformer layers. We denote the sequence of layers as $F = (F_0, F_1, \dots, F_{L-1})$. 
Concretely, given newly generated token $y_{p,q}$, both branches share a unified backbone of depth $m$:
\begin{equation}
h_{p,q} = F_{0:m-1}(y_{p,q}),
\label{eq:shared-trunk}
\end{equation}
From this shared intermediate state, we construct two distinct pathways:
\begin{align}
z^{H}_{p,q} &= Head^H(F_{m:L-1}\bigl(h_{p,q})), \label{eq:h-branch}\\
z^{V}_{p,q} &= Head^V(\widetilde{F}_{m:L-1}\bigl(h_{p,q})). \label{eq:v-branch}
\end{align}
Here, the horizontal branch utilizes the original top $L-m$ layers $F_{m:L-1}$ in the pre-trained model, whereas the vertical branch employs $\widetilde{F}_{m:L-1}$, an independently trainable block initialized as a clone of the top $L-m$ layers. The original decoding head $Head^H$ is attached to the horizontal branch, serving for row-wise prediction. A complementary vertical decoding head $Head^V$ is attached to the vertical branch for column-wise prediction. 
%


Compared to NAR~\cite{nar}, this branching design also improves runtime efficiency. Attaching a block after final layer would extend the critical-path depth. In contrast, branching at depth $m$ allows the horizontal and vertical blocks to execute concurrently on top of the shared trunk, effectively maintaining the original critical-path depth.

\subsection{Learnable Fusion Gate}
The horizontal and vertical heads provide two complementary predictions for each image token. 
%
%
While simple averaging adopted by NAR~\cite{nar} assumes an isotropic contribution from both horizontal and vertical contexts, natural images often exhibit significant anisotropy. For instance, when predicting a pixel along a sharp horizontal edge, the vertical predecessor may provide more reliable structural continuity than the horizontal one. A static average fails to capture such directional dependencies.
Moreover, when the horizontal and vertical heads yield conflicting logit distributions, direct averaging acts as a low-pass filter in the probability space, which may lead to blurred textures or artifacts.
We therefore fuse the two predictions with a learnable fusion gate, rather than simply averaging them.

Specifically, for $p>0$ and $q>0$, we compute a context-dependent gate from the two predecessor states:
\begin{equation}
g_{p,q}=\sigma\!\left(\mathrm{MLP}\!\left([\,h^{H}_{p,q-1};\,h^{V}_{p-1,q}\,]\right)\right),
\label{eq:interior-gate}
\end{equation}
where $[\,\cdot;\cdot\,]$ denotes concatenation and $\sigma$ is the sigmoid function. 
The fused logit is then given by
\begin{equation}
z_{p,q}=g_{p,q}\, z^{H}_{p,q-1}
+(1-g_{p,q})\, z^{V}_{p-1,q}.
\label{eq:interior-fusion}
\end{equation}
Notably, for boundary positions, only the available directional prediction is used: the first row is decoded by horizontal logits, and the first column is decoded by vertical logits. The corner token $(0,0)$ is predicted from the conditioning prefix.

\subsection{Stabilized Adaptation and Training Objectives} \label{sec:two_stage}

\textbf{Two-stage stabilized adaptation.} To retain the generative capabilities of the pre-trained AR model while seamlessly adapting it for parallel decoding, we adopt a stabilized, two-stage post-training paradigm:
\begin{itemize} [leftmargin=*]
    \item \textit{Stage 1: Branch Initialization.} In the initial phase, we completely freeze the pre-trained transformer backbone along with the original horizontal head. Only the newly initialized vertical head and the dynamic gating mechanism are set as trainable. This frozen-backbone strategy effectively prevents catastrophic forgetting of the pre-trained visual manifold and ensures highly stable, rapid convergence for the new components.
    \item \textit{Stage 2: Joint Full Fine-Tuning.} Once the vertical pathway is sufficiently aligned, we unfreeze the entire network. The backbone, horizontal head, vertical head, and gating module are jointly fine-tuned. This stage harmonizes the orthogonal spatial representations and fully optimizes the network for 2D diagonal-parallel generation.
\end{itemize}



\textbf{Training objectives.} To optimize the model across these stages, we employ a comprehensive objective function comprising fusion and auxiliary losses. We supervise the final gated logits ($z_{p,q}$) via the standard Cross-Entropy (CE) loss to ensure target fidelity:
\begin{equation}
\mathcal{L}_{\text{fuse}} = \frac{1}{HW} \sum_{p=0}^{H-1}\sum_{q=0}^{W-1} \mathrm{CE}\!\left(z_{p,q},\, y_{p,q}\right).
\label{eq:loss-fuse}
\end{equation}
Furthermore, to ensure the standalone predictive capability of both the horizontal and vertical pathways under dynamic gating, we apply auxiliary CE losses to their independent predictions:
\begin{equation}
\begin{aligned}
\mathcal{L}_{H}
&= \frac{1}{H(W-1)}
\sum_{p=0}^{H-1}\sum_{q=0}^{W-2}
\mathrm{CE}\!\left(z^{H}_{p,q},\, y_{p,q+1}\right), \\
\mathcal{L}_{V}
&= \frac{1}{(H-1)W}
\sum_{p=0}^{H-2}\sum_{q=0}^{W-1}
\mathrm{CE}\!\left(z^{V}_{p,q},\, y_{p+1,q}\right).
\end{aligned}
\end{equation}

Finally, the overall training objective is defined as: $\mathcal{L} = \mathcal{L}_{\text{fuse}} + \lambda_{\text{aux}}(\mathcal{L}_{H} + \mathcal{L}_{V})$.

\subsection{Inference-Time Parallelization with KV Cache} \label{sec:inference_optim}

During inference, image generation proceeds iteratively along diagonals $\mathcal{D}_t$ in strict parallel, where logits for $\mathcal{D}_t$ are fused directly from the preceding diagonal $\mathcal{D}_{t-1}$. To translate this theoretical parallelism into inference speedups, we implement a hardware-aware inference pipeline. First, to enforce our 2D diagonal-step masking without the severe memory fragmentation caused by dense zero-padding, we integrate FlexAttention~\cite{dong2024flexattentionprogrammingmodel}. This optimization dynamically compiles our highly sparse proximity mask on the fly, entirely bypassing the materialization of the full attention matrix. Second, after concurrently sampling $\mathcal{D}_t$, all tokens are appended to the branch-specific KV caches in a single batched operation. We similarly batch the conditional and unconditional forward passes for classifier-free guidance~\cite{ho2022classifierfreediffusionguidance}, effectively amortizing kernel launch overheads and maximizing GPU arithmetic intensity.





\section{Experiments}

\subsection{Experimental Setup}

\textbf{Training details.}
Our main experiments use LlamaGen~\cite{llamagen} for ImageNet $256\times256$ class-conditional image generation, and Emu3.5-Image-34B~\cite{cui2025emu35} for text-to-image synthesis at $512\times512$ resolution. For LlamaGen, post-training is conducted for 25 epochs on the ImageNet dataset~\cite{russakovsky2015imagenetlargescalevisual} with a batch size of 512. For Emu3.5-Image-34B, we curate a compact corpus of approximately 80K text-image pairs from OpenGPT-4o-Image~\cite{chen2025opengpt4oimagecomprehensivedatasetadvanced} and ShareGPT-4o-Image~\cite{chen2025sharegpt4oimagealigningmultimodalmodels}, running the post-training stage for 50K steps. The additional GLM-Image~\cite{glmimage2026} experiment uses 40M post-training tokens.

As detailed in Section~\ref{sec:two_stage}, we adopt a two-stage training schedule. Stage 1 freezes the backbone during the first 20\% of the training steps to stabilize the vertical head, whereas stage 2 unlocks the full network and scales the backbone learning rate by a factor of 0.2 relative to the head. Both stages use a base learning rate of $2\times10^{-5}$ with cosine decay. Loss coefficients are $\lambda_{\text{aux}}=0.05$. Classifier-free guidance (cfg) is set to $2.0$ for LlamaGen and $5.0$ for Emu3.5 by default. All experiments are run on 8 NVIDIA H20 GPUs with bf16 precision.

\textbf{Baselines and metrics.}
We compare FlashAR against three categories of baselines. (i) Raster-scan AR: the original LlamaGen, Emu3.5-Image, and GLM-Image checkpoints with sequential raster-scan decoding. (ii) Parallel AR trained from scratch: VAR~\cite{tian2024var}, PAR~\cite{wang2025par}, 
and NAR~\cite{nar}, which adopt distinct parallelization strategies but all 
require training from scratch under modified generation objectives. (iii) Post-training adaptation: Block Diffusion, which is adopted by Emu3.5~\cite{cui2025emu35} and enables parallel generation via discrete diffusion adaptation. For LlamaGen, we report FID~\cite{heusel2018ganstrainedtimescaleupdate}, Inception Score (IS)~\cite{salimans2016improvedtechniquestraininggans}, Precision and Recall~\cite{ghosh2023genevalobjectfocusedframeworkevaluating}, and sFID as generative quality metrics. For text-to-image generation, we report GenEval~\cite{ghosh2023genevalobjectfocusedframeworkevaluating} on Emu3.5-Image and GLM-Image. On Emu3.5-Image, we additionally use DPG-Bench~\cite{hu2024ella}, T2I-CompBench~\cite{huang2023compbench}, and PickScore~\cite{kirstain2023pickapic} to assess dense-prompt alignment, fine-grained composition, and predicted human preference, respectively. Inference latency is measured on a single NVIDIA H20-96G GPU with batch size 1, averaged over 100 samples.

\subsection{Main Results}

\begin{table*}[t]
  \centering
  \small
  \caption{Quantitative evaluation on the ImageNet 256$\times$256 benchmark}
  \label{tab:main_compare}
  \resizebox{\linewidth}{!}{
  \begin{tabular}{clccccccc}
    \toprule
    Model size & Method & Type & Training Epoch & FID$\downarrow$ & IS$\uparrow$ & P/R-F1$\uparrow$ & Steps & \begin{tabular}[c]{@{}c@{}}Throughput\\ (img/s)\end{tabular} \\
    \midrule
    \multirow{5}{*}{\begin{tabular}[c]{@{}c@{}}B \\ ($\sim$120M)\end{tabular}}
      & LlamaGen      & From scratch  & 300 & 5.46          & 193.6          & 0.594 & 256 & 117.9          \\
      & PAR           & From scratch  & 300 & 6.21          & 204.4          & 0.537 & 67  & 174.1          \\
      & NAR           & From scratch  & 300 & \textbf{4.65}          & \textbf{212.3}          & 0.600 & 31  & 419.7          \\
      & BlockDiffusion & Post-training & 75  & 5.91 & 176.2 & 0.589 & 64 & 186.3           \\
      & FlashAR       & Post-training & 25  & 4.68          & 208.3          & \textbf{0.605} & 31  & \textbf{447.2} \\
    \midrule
    \multirow{6}{*}{\begin{tabular}[c]{@{}c@{}}L \\ ($\sim$360M)\end{tabular}}
      & LlamaGen      & From scratch  & 300 & 3.80          & 248.3          & 0.639 & 256 & 47.1           \\
      & PAR           & From scratch  & 300 & 4.32          & 189.4          & 0.576 & 67  & 93.8           \\
      & NAR           & From scratch  & 300 & \textbf{3.06}          & 263.9          & 0.641 & 31  & 195.4          \\
      & VAR           & From scratch  & 200 & 3.30          & 274.4          & 0.634 & 10  & 129.3          \\
      & BlockDiffusion & Post-training & 75  & 4.55 & 243.5 & 0.645 & 64 & 103.2          \\
      & FlashAR       & Post-training & 25  & 3.16          & \textbf{289.0}          & \textbf{0.656} & 31  & \textbf{224.7} \\
    \midrule
    \multirow{5}{*}{\begin{tabular}[c]{@{}c@{}}XL \\ ($\sim$700M)\end{tabular}}
      & LlamaGen      & From scratch  & 300 & 3.39          & 227.1          & 0.648 & 256 & 23.7           \\
      & PAR           & From scratch  & 300 & 3.50          & 234.4          & 0.619 & 67  & 53.9           \\
      & NAR           & From scratch  & 300 & \textbf{2.70}          & 277.5          & \textbf{0.676} & 31  & 98.1           \\
      & BlockDiffusion & Post-training & 75  & 4.13 & 258.6 & 0.654 & 64 & 41.7            \\
      & FlashAR       & Post-training & 25  & 2.94            & \textbf{293.7}	             & 0.672    & 31  & \textbf{109.3}             \\
    \midrule
    \multirow{5}{*}{\begin{tabular}[c]{@{}c@{}}XXL \\ ($\sim$1.4B)\end{tabular}}
      & LlamaGen      & From scratch  & 300   & 3.09          & 253.6          & 0.647 & 256 & 14.1           \\
      & PAR           & From scratch  & 300   & 3.20          & 288.3          & 0.632 & 67  & 33.9           \\
      & NAR           & From scratch  & 300   & \textbf{2.58}          & \textbf{293.5}          & 0.673 & 31  & 56.9           \\
      & BlockDiffusion & Post-training & 75  & 3.78 & 264.9 & 0.652 & 64 & 26.8            \\
      & FlashAR       & Post-training & 25  & 2.79            & 289.4            & \textbf{0.690}   & 31  & \textbf{63.4}             \\
    \bottomrule
  \end{tabular}
  }
\end{table*}

\textbf{Results on LlamaGen.}
Table~\ref{tab:main_compare} presents the class-conditional image generation results on ImageNet at $256\times256$ resolution. Among existing post-training methods, FlashAR significantly outperforms BlockDiffusion in both quality and efficiency. Specifically, at the L scale, FlashAR achieves a superior FID (3.16 vs. 4.55) using only 25 adaptation epochs---one-third of BlockDiffusion's training budget. More notably, FlashAR-L obtains an IS that surpasses NAR-L (289.0 vs. 263.9), a model trained entirely from scratch, despite FlashAR requiring only lightweight post-training on a pre-trained raster-AR backbone. Additionally, FlashAR-B achieves a throughput of \textbf{447.21}~images/s, outperforming even NAR-B (419.7~images/s); this efficiency gain is attributable to the inference-time optimizations detailed in Section~\ref{sec:inference_optim}. Overall, these results demonstrate that the proposed post-training approach yields both competitive generation quality and superior inference efficiency across various model scales.


\textbf{Results on Emu3.5-Image.}
We further evaluate FlashAR on Emu3.5-Image-34B~\cite{cui2025emu35} for $512\times512$ text-to-image generation. FlashAR retains competitive generation quality, achieving an overall GenEval score of 80.17, compared with 80.48 for the original model and 74.22 for our BlockDiffusion reproduction (Table~\ref{tab:emu35_geneval}).  Additional text-to-image evaluations in Table~\ref{tab:broader_t2i} show our method achieves comparable performance with the original model in DPG-Bench and PickScore, alongside improvements in four T2I-CompBench dimensions and modest decreases in shape and texture.

Alongside this competitive quality, FlashAR substantially accelerates generation. After 50K post-training steps using 80M tokens---approximately 0.053\% of the backbone's reported pre-training tokens---FlashAR reduces the number of sequential decoding steps from 1024 to 63. As shown in Table~\ref{tab:emu35}, latency decreases from 103.57 to 7.95 seconds per image, yielding a $13.03\times$ speedup over the fastest evaluated AR baseline, with batched CFG enabled for both methods.

\begin{table*}[h]
  \centering
  \small
  \caption{GenEval scores on Emu3.5-Image-34B (cfg\,=\,5.0, $512\times512$).}
  \label{tab:emu35_geneval}
  \resizebox{0.92\linewidth}{!}{
  \begin{tabular}{@{}l c c c c c c c@{}}
    \toprule
    {\textbf{Method}}& \textbf{Overall} & \textbf{Single Obj} & \textbf{Two Obj} & \textbf{Counting} & \textbf{Colors} & \textbf{Position} & \textbf{Color\_attr} \\
    \midrule
    Emu3.5-Image & 80.48 & 100.00 & 94.95 & 53.75 & 90.96 & 73.00 & 70.25 \\
    BlockDiffusion & 74.22   & 96.88 & 88.89 & 47.50 & 85.64 & 68.00 & 58.44 \\
    FlashAR & 80.17 & 98.75 & 91.92 & 53.75 & 92.55 & 80.00 & 64.00 \\
    \bottomrule
  \end{tabular}
  }
\end{table*}

\begin{table}[ht]
\centering\small
\caption{Text-to-image generation results on Emu3.5-Image.}
\label{tab:broader_t2i}
\setlength{\tabcolsep}{4pt}
\resizebox{\linewidth}{!}{%
\begin{tabular}{lrrrrrrrrr}
\toprule
\multirow{2}{*}{Method} & \multirow{2}{*}{GenEval $\uparrow$} & \multirow{2}{*}{DPG-Bench $\uparrow$} & \multirow{2}{*}{PickScore $\uparrow$} & \multicolumn{6}{c}{T2I-CompBench} \\
\cmidrule(lr){5-10}
 & & & & Non-spatial $\uparrow$ & Complex $\uparrow$ & Color $\uparrow$ & Shape $\uparrow$ & Texture $\uparrow$ & Spatial $\uparrow$ \\
\midrule
Original AR & 80.48 & 86.74 & 0.2309 & 0.31 & 0.30 & 0.81 & 0.63 & 0.75 & 0.41 \\
FlashAR & 80.29 & 86.12 & 0.2262 & 0.32 & 0.32 & 0.82 & 0.61 & 0.73 & 0.42 \\
\bottomrule
\end{tabular}
}
\par
\end{table}

\begin{table*}[ht]
  \centering
  \small
  \caption{Inference efficiency on Emu3.5-Image ($512\times512$, CFG$=5.0$). Speedups are relative to the fastest evaluated AR baseline; both it and FlashAR use batched CFG.}
  \label{tab:emu35}
  \begin{tabular}{@{}llccc@{}}
    \toprule
    Method & Attention backend & Batched CFG & Latency (s) & Speedup \\
    \midrule
    Original AR & FlashAttention-2 & No & 189.36 & $0.55\times$ \\
    Original AR & FlexAttention & Yes & 121.83 & $0.85\times$ \\
    Original AR & FlashAttention-2 & Yes & 103.57 & $1.00\times$ \\
    FlashAR & FlexAttention & Yes & 7.95 & $13.03\times$ \\
    \bottomrule
  \end{tabular}
\end{table*}

\textbf{Results on GLM-Image.}
As shown in Table~\ref{tab:extra_backbones}, beyond Emu3.5-Image, we apply the same adaptation framework to
GLM-Image~\cite{glmimage2026}. With 40M post-training tokens,
FlashAR reduces latency from 110.17 to 8.92 seconds per image
($12.35\times$), while GenEval decreases from 70.34 to 68.97.
\par

\begin{table}[ht]
\centering\small
\caption{Text-to-image generation results on GLM-Image. Speedup is measured relative to the original AR implementation used in this experiment.}
\label{tab:extra_backbones}
\setlength{\tabcolsep}{4pt}
\resizebox{\linewidth}{!}{%
\begin{tabular}{lrrrrrrrrr}
\toprule
\multirow{2}{*}{Method} & \multirow{2}{*}{Latency (s) $\downarrow$} & \multirow{2}{*}{Speedup} & \multicolumn{7}{c}{GenEval $\uparrow$} \\
\cmidrule(lr){4-10}
 & & & Overall & Single Obj & Two Obj & Counting & Colors & Position & Color Attr. \\
\midrule
Original AR & 110.17 & $1.00\times$ & 70.34 & 95.14 & 81.73 & 45.29 & 88.16 & 64.82 & 44.05 \\
FlashAR & 8.92 & $12.35\times$ & 68.97 & 94.62 & 79.84 & 44.81 & 87.42 & 61.35 & 43.19 \\
\bottomrule
\end{tabular}
}
\par
\end{table}

\FloatBarrier
\subsection{Ablation Study}


\begin{figure}[H]
  \centering
  \begin{minipage}[t]{0.43\textwidth}
    \centering
\begin{tikzpicture}
  \begin{axis}[
    width=\linewidth,
    height=5.5cm,
    title={(a) Two-stage Fine-tuning Improves Convergence},
    xlabel={Training Epochs},
    ylabel={FID $\downarrow$},
    xmin=0, xmax=25,
    ymin=3.0, ymax=6.0,
    xtick={0,5,10,15,20,25},
    ytick={3.0,3.6,4.2,4.8,5.4,6.0},
    grid=both,
    major grid style={line width=0.2pt, draw=black!15},
    minor grid style={line width=0.1pt, draw=black!8},
    minor tick num=1,
    tick label style={font=\small},
    label style={font=\small},
    title style={font=\small},
    legend style={
      at={(0.5,-0.18)},
      anchor=north,
      legend columns=3,
      font=\scriptsize,
      fill=none,
      draw=none
    },
    clip=false
  ]
    \addplot+[
      very thick,
      color=black!70,
      mark=square*,
      mark size=1.8pt,
      mark options={fill=black!70},
      smooth
    ] coordinates {
      (5, 4.42)
      (10, 3.90)
      (15, 3.56)
      (20, 3.33)
      (25, 3.27)
    };
    \addlegendentry{One-stage}

    \addplot+[
      very thick,
      color=blue!85!black,
      mark=triangle*,
      mark size=2.0pt,
      mark options={fill=blue!85!black},
      smooth
    ] coordinates {
      (5, 3.98)
      (10, 3.50)
      (15, 3.33)
      (20, 3.23)
      (25, 3.16)
    };
    \addlegendentry{Two-stage}

    \addplot+[
      very thick,
      color=red!70!black,
      mark=*,
      mark size=1.8pt,
      mark options={fill=red!70!black},
      smooth
    ] coordinates {
      (5, 5.62)
      (10, 5.08)
      (15, 4.82)
      (20, 4.65)
      (25, 4.55)
    };
    \addlegendentry{BlockDiffusion}

    \addplot[
      blue!55!black,
      densely dashed,
      line width=0.7pt
    ] coordinates {
      (5, 3.0)
      (5, 6.0)
    };

    \node[
      anchor=north,
      font=\scriptsize,
      text=blue!60!black
    ] at (axis cs:2.5,5.95) {Stage 1};

    \node[
      anchor=north,
      font=\scriptsize,
      text=blue!60!black
    ] at (axis cs:17.5,5.95) {Stage 2};

  \end{axis}
\end{tikzpicture}
  \end{minipage}
  \hfill
  \begin{minipage}[t]{0.5\textwidth}
    \centering
    \begin{tikzpicture}
      \begin{axis}[
        width=\linewidth,
        height=5.5cm,
        title={(b) Component Ablation (cfg=1.5)},
        ybar,
        bar width=16pt,
        ylabel={Final FID $\downarrow$},
        ymin=3.6, ymax=4.6,
        symbolic x coords={Naive Adaptation,+Upper Branch, +Fusion Gate,FlashAR},
        xtick=data,
        x tick label style={rotate=15,anchor=east,font=\small},
        tick label style={font=\small},
        label style={font=\small},
        nodes near coords,
        every node near coord/.append style={font=\small},
        major grid style={line width=0.2pt, draw=black!30},
        ymajorgrids=true,
        axis on top
      ]
        \addplot[fill=blue!60, draw=blue!90] coordinates {(Naive Adaptation, 4.36) (+Upper Branch,4.16) (+Fusion Gate,4.12) (FlashAR,3.98)};
      \end{axis}
    \end{tikzpicture}
  \end{minipage}
  \caption{Ablation studies on LlamaGen-L. \textbf{(a)} FID convergence trajectories across training epochs at CFG$=2.0$. \textbf{(b)} Final FID comparisons across component variants at CFG$=1.5$.}
  \label{fig:ablation}
\end{figure}

\textbf{Post-training efficiency comparison with block diffusion.}
To evaluate the convergence efficiency of our approach, we plot the FID trajectories across training epochs in Figure~\ref{fig:ablation}(a). As illustrated, FlashAR demonstrates significantly faster convergence and superior generation quality compared to the BlockDiffusion baseline. Remarkably, after only 5 epochs of adaptation, the two-stage FlashAR achieves an FID of 3.98, which already comfortably outperforms the final performance of BlockDiffusion at 25 epochs (4.55). 
Furthermore, the results validate the efficacy of our proposed two-stage fine-tuning schedule over the conventional one-stage baseline, where the two-stage schedule ultimately reaches a superior FID of 3.16 compared to 3.27 for the one-stage baseline at epoch 25.

\textbf{Effect of branching depth.}
Table~\ref{tab:branch_depth} evaluates the impact of the branching depth $m$. Excessive branching depth limits the expressive capacity of the vertical pathway, whereas insufficient depth reduces trunk sharing and increases parameter overhead. We find that pre-final branching ($m=23$ of $L=24$ layers) yields the optimal trade-off. This observation is consistent with the component ablation results presented in Figure~\ref{fig:ablation}(b) under a different cfg scale. Specifically, introducing the vertical branch alone improves the FID over naive adaptation (4.16 vs.\ 4.36), validating our design choice to branch from pre-final layers rather than relying solely on the final-layer representation.

\begin{table}[ht]
  \centering
  \small
  \caption{Effect of branching depth $m$ on LlamaGen-L ($L=\text{24}$, cfg$=2.0$).}
  \label{tab:branch_depth}
  \setlength{\tabcolsep}{5pt}
  \begin{tabular}{cccccc}
    \toprule
    \textbf{Branch Depth} & \textbf{FID} $\downarrow$ & \textbf{sFID} $\downarrow$ & \textbf{IS} $\uparrow$ & \textbf{Prec.} $\uparrow$ & \textbf{Recall} $\uparrow$ \\
    \midrule
    $m=24$        & 3.29          & 6.41          & 288.0          & \textbf{0.835} & 0.529 \\
    $m=23$ (ours) & \textbf{3.16} & 6.34 & \textbf{289.0} & 0.834          & \textbf{0.535} \\
    $m=21$        & 3.20          & \textbf{6.32}          & 285.9          & 0.833          & 0.518 \\
    \bottomrule
  \end{tabular}
\end{table}

\textbf{Effect of fusion gate design.}
Table~\ref{tab:gate_ablation} compares various strategies for combining horizontal and vertical predictions. Fixed averaging is consistently inferior to our learnable fusion gate in terms of FID, sFID, and IS, indicating that the relative importance of the two directional cues should be determined adaptively rather than prescribed globally. The component ablation in Figure~\ref{fig:ablation}(b) under a different CFG scale further demonstrates the efficacy of the learnable fusion gate, which reduces the FID from 4.36 to 4.12.

\begin{table}[h]
  \centering
  \small
  \caption{Ablation on fusion strategies (cfg=1.5).}
  \label{tab:gate_ablation}
  \begin{tabular}{lccccc}
    \toprule
    \textbf{Fusion Strategy} & \textbf{FID} $\downarrow$ & \textbf{sFID} $\downarrow$ & \textbf{IS} $\uparrow$ & \textbf{Prec.} $\uparrow$ & \textbf{Recall} $\uparrow$ \\
    \midrule
    Average ($g=0.5$)     & 4.36 & 6.32 & 187.8 & 0.756 & \textbf{0.604} \\
    Learnable Gate (ours) & \textbf{4.12} & \textbf{6.30} & \textbf{191.4} & \textbf{0.765} & 0.596 \\
    \bottomrule
  \end{tabular}
\end{table}



\FloatBarrier
\section{Conclusion}

In this paper, we have proposed FlashAR, a lightweight post-training adaptation framework that efficiently transforms a pre-trained raster-scan autoregressive model into a highly parallel generator. To bypass the inherent horizontal bias of the pre-trained model and mitigate context conflicts, we have branched the vertical pathway from a pre-final layer and have integrated a learnable fusion gate to dynamically combine directional predictions at each target position.
Furthermore, we have designed a two-stage post-training pipeline that minimizes adaptation overhead through vertical branch initialization followed by joint fine-tuning. 
On the hardware deployment front, we have implemented a hardware-aware inference pipeline leveraging FlexAttention and batched KV-cache updates. 
Extensive experiments have demonstrated a 13.03$\times$ speedup on Emu3.5-Image using approximately 0.05\% of its pre-training tokens, while retaining competitive generation quality.
%

\paragraph{Limitations and future work.} 
Despite its efficacy, FlashAR has two primary limitations. First, determining the optimal pre-final branching depth currently relies on empirical search. Second, generating tokens on the same anti-diagonal in strict parallel enforces conditional independence, depriving them of intra-diagonal mutual perception.
In future work, we will explore automated architecture search for branch placement and lightweight intra-step communication (e.g., iterative refinement) to restore mutual perception. Additionally, we plan to extend this diagonal-parallel paradigm to 3D spatio-temporal tasks, such as video autoregressive decoding.


\bibliographystyle{iclr2026_conference}
{
\small
\bibliography{reference}
}

\newpage
\appendix

\section{Visualization}

\begin{figure}[h]
    \centering
    \includegraphics[width=0.975\linewidth]{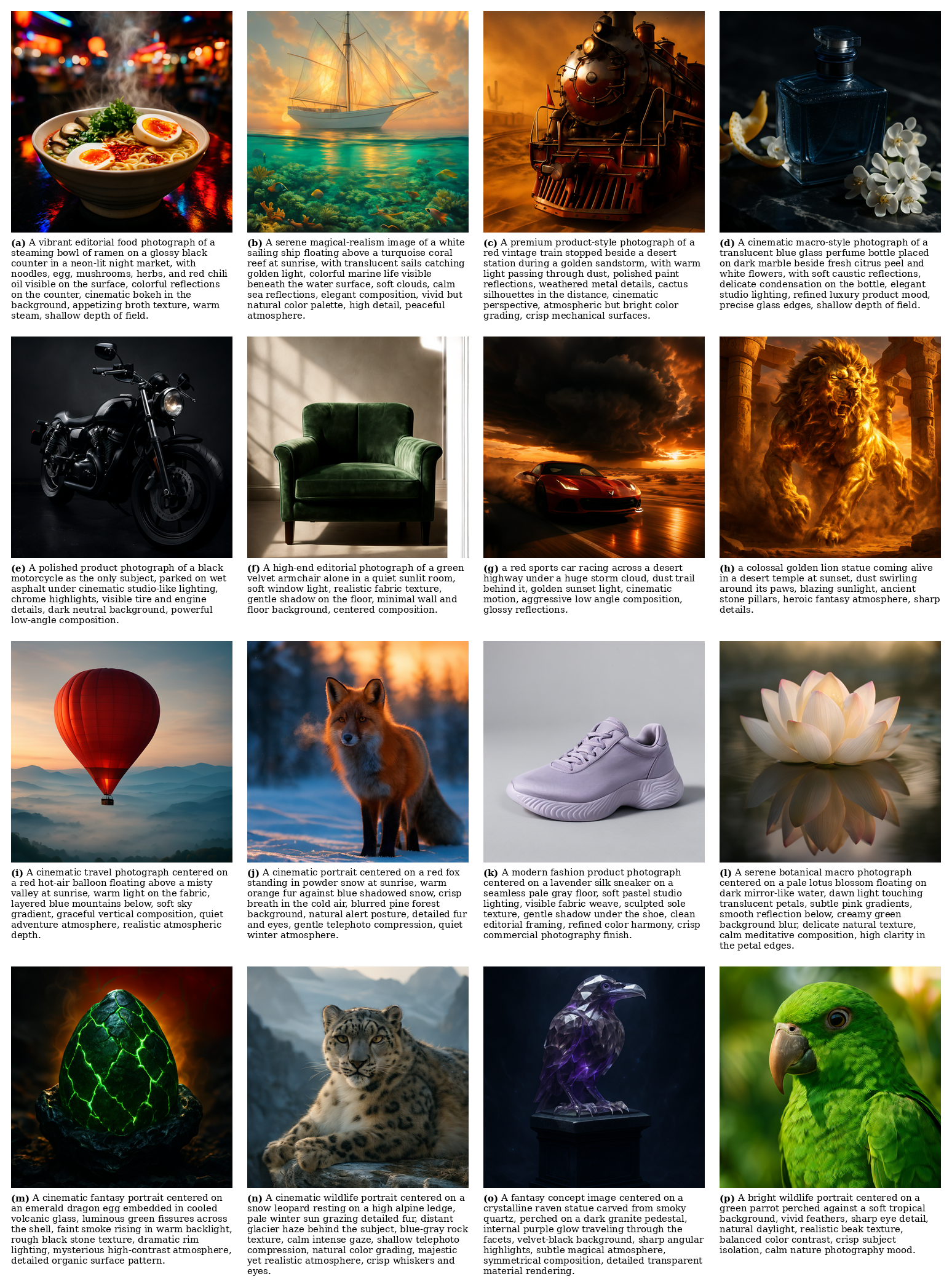}
    \caption{\textbf{Complex text-guided image generation samples} by Emu3.5-Image-FlashAR}
    \label{fig:placeholder}
\end{figure}

\begin{figure}[p]
    \centering
    \begin{minipage}{\linewidth}
        \centering
        \includegraphics[width=1\linewidth]{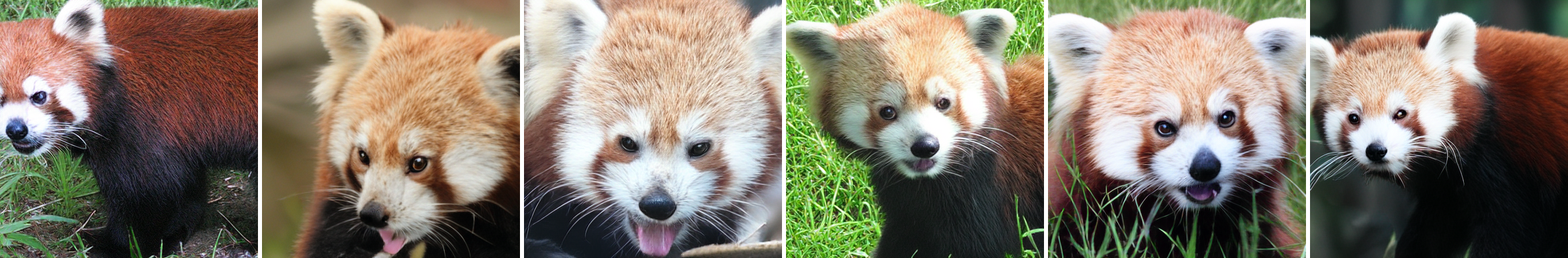}
        \caption*{class id 387, lesser panda}
    \end{minipage}
    \vspace{0.8em}
    \begin{minipage}{\linewidth}
        \centering
        \includegraphics[width=\linewidth]{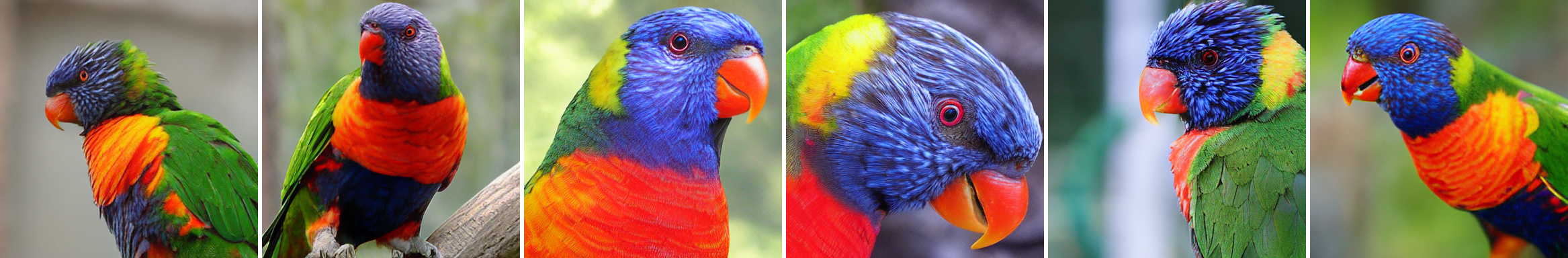}
        \caption*{class id 90, lorikeet}
    \end{minipage}
    \hfill
    \begin{minipage}{\linewidth}
        \centering
        \includegraphics[width=\linewidth]{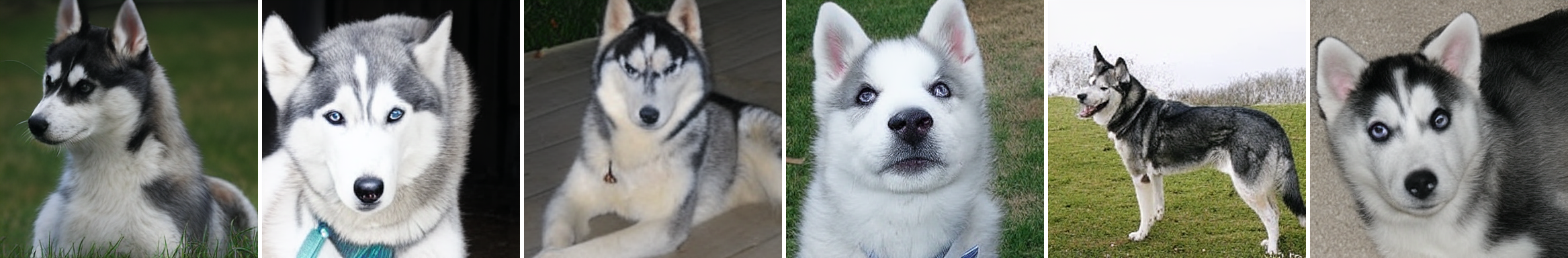}
        \caption*{class id 250, Siberian husky}
    \end{minipage}

    \vspace{0.8em}
    \begin{minipage}{\linewidth}
        \centering
        \includegraphics[width=1\linewidth]{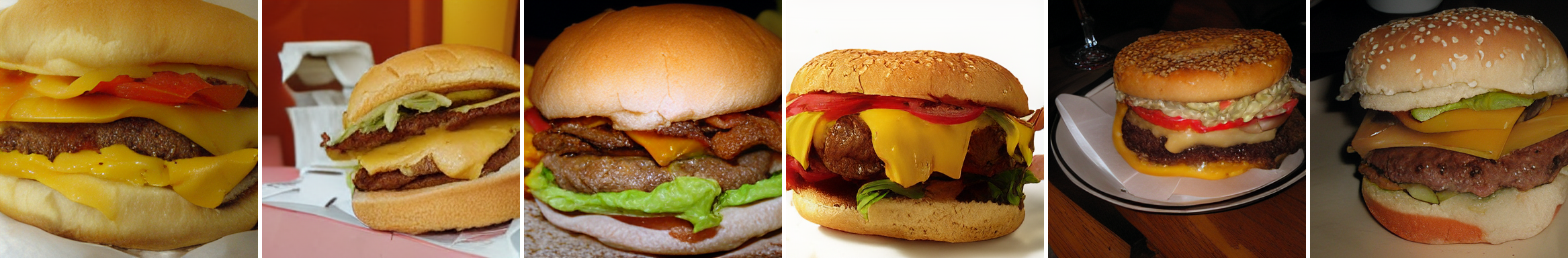}
        \caption*{class id 933, cheeseburger}
    \end{minipage}
    
    \begin{minipage}{\linewidth}
        \centering
        \includegraphics[width=\linewidth]{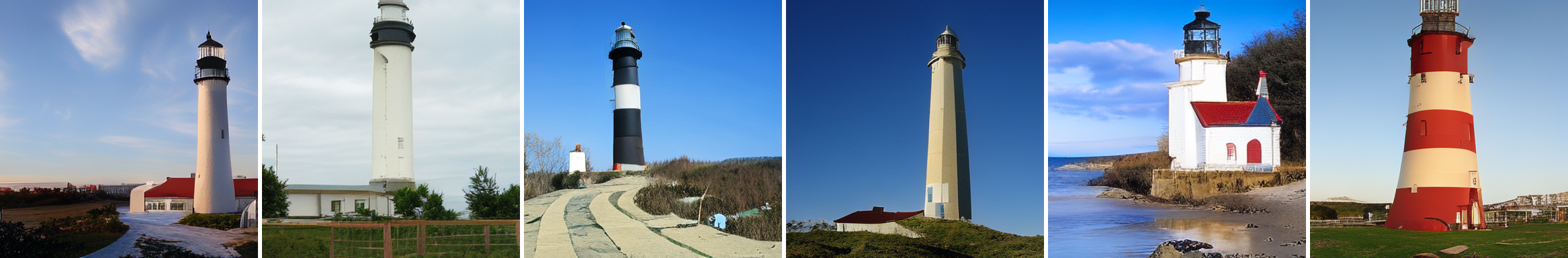}
        \caption*{class id 437, beacon}
    \end{minipage}
    
    \hfill
    \begin{minipage}{\linewidth}
        \centering
        \includegraphics[width=\linewidth]{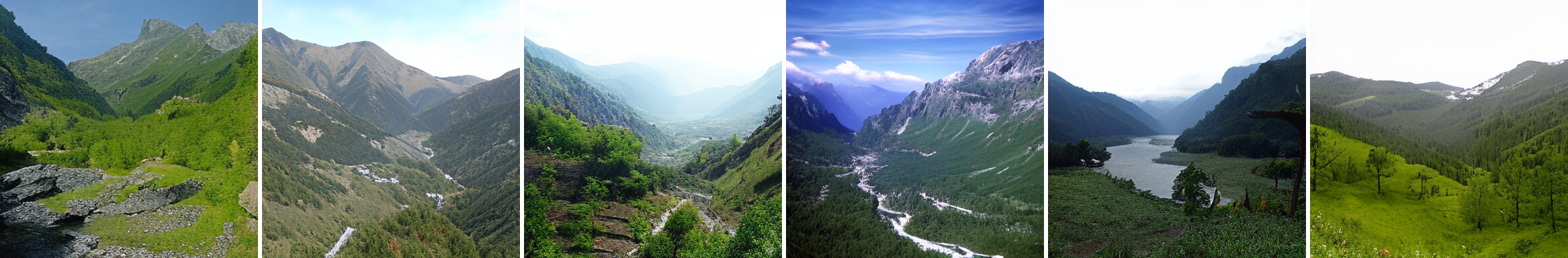}
        \caption*{class id 979, valley}
    \end{minipage}

    \vspace{0.8em}

    \begin{minipage}{\linewidth}
        \centering
        \includegraphics[width=\linewidth]{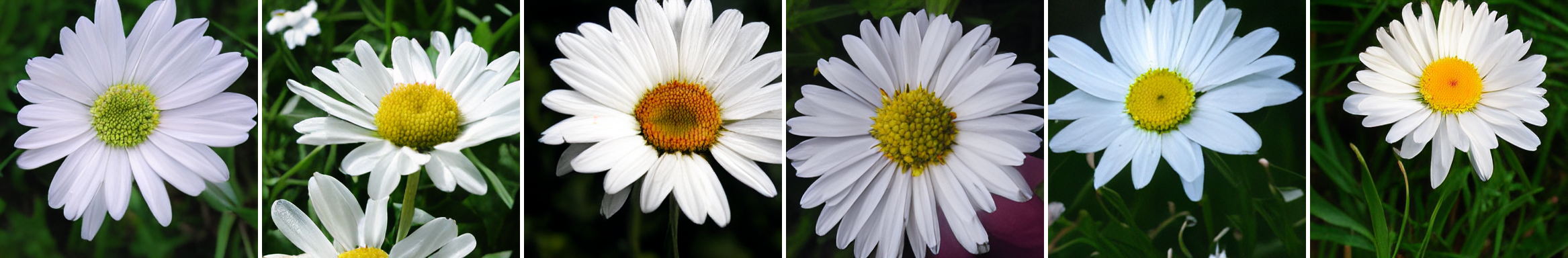}
        \caption*{class id 985, daisy}
    \end{minipage}

    \caption{ \textbf{Class-conditional image generation samples} produced by FlashAR-XXL on \text{Imagenet} $256 \times 256$ }
    \label{fig:visualization}
\end{figure}


\end{document}